# *Human vs. Teleoperated Robots in Vineyard Management: A Simulation-Based Analysis of Travel Speed, Routing, and Task Performance*

Daniel Udekwe, Hasan Seyyedhasani, Muhammad Ali Qadri
School of Plant and Environmental Sciences, Virginia Tech, Blacksburg VA-24060, USA

Email of corresponding authors:
daudekwe@vt.edu (Daniel Udekwe)
seyyedhasani12@vt.edu (Hasan Seyyedhasani)
muhammadaliq@vt.edu (Muhammad Ali Qadri)

# Abstract

Rising labor costs and narrow treatment windows have made teleoperated robots a proposed tool for vineyard scouting and treatment, but simulation-based comparisons of human and robotic performance are sensitive to modeling choices that can favor one agent before any trial runs: whether both can record a target's location, whether both are routed efficiently, and whether travel speeds are assessed consistently. This paper simulates a human operator against a teleoperated robot on two vineyard tasks, disease detection-and-treatment and yield-zone survey, holding both agents to matched location-marking and routing standards, using a routing strategy proven optimal for this topology, and modeling teleoperation-specific effects. Once information access and routing are held equal, task-time outcomes are governed largely by relative sustained travel speed: at the speeds modeled (1.4 m/s human, 0.9 m/s UGV), the human completes both tasks 43-48 percent faster, a gap persisting across a simulated eight-cycle season, while at speed parity the agents differ by only 9-12 percent, and every other parameter shifts the result by under 4 points. These results indicate the comparison is primarily a question of platform speed, and that the case for teleoperated robots should rest on labor availability, endurance, consistency, or hazard avoidance rather than completion time..

**Keywords:** Human-robot interaction; Precision viticulture; Teleoperation; Field robotics; Simulation; Route optimization

# 1. Introduction

Vineyard management is labor-intensive, time-sensitive, and increasingly constrained by labor availability and operating costs. Disease scouting and treatment require timely field coverage because delays in identifying and addressing disease can allow localized outbreaks to expand rapidly (Giffard et al., 2022; Jogaiah, 2023). Similarly, yield-map-informed interventions must be performed within crop-specific phenological windows, such as pre-bloom and veraison, to be effective (L. F. P. Oliveira et al., 2021). These operational demands create a need for management strategies that can maintain timely field coverage while reducing dependence on manual labor (Goel et al., 2021). Agricultural ground robots equipped with sensing and manipulation capabilities have therefore emerged as a potential means of automating repetitive scouting, monitoring, and treatment activities while allowing human workers to focus on tasks requiring higher-level judgment (Conejero et al., 2025; Massah et al., 2021; Mohimont et al., 2022).

Teleoperation provides one pathway for deploying such robotic systems when fully autonomous operation remains impractical (Chen et al., 2020a; Udekwe & Seyyedhasani, 2025a). In particular, virtual reality (VR) interfaces can provide operators with depth perception, spatial context, and an immersive view of the remote environment that may be advantageous for tasks requiring three-dimensional spatial judgment (Udekwe & Seyyedhasani, 2025a). This capability is potentially relevant to vineyard operations, where dense canopies and occluded targets can make disease identification and manipulation difficult (T. Liu et al., 2024a; Udekwe & Seyyedhasani, 2025c; Wang et al., 2024). However, the presence of an immersive interface does not, by itself, establish an operational advantage. Any benefit of VR-based teleoperation must ultimately be evaluated in terms of task performance, including detection, travel, treatment, and overall completion time.

A fundamental challenge in evaluating human and robotic field operations is ensuring that the comparison itself is fair. Simulation provides an effective means of conducting such comparisons because it permits repeated experiments while controlling field geometry, target distributions, movement speeds, routing strategies, and task characteristics. However, simulation-based comparisons can produce misleading conclusions when seemingly reasonable modeling choices systematically favor one agent. For example, assigning persistent target-location memory to a robot gives the robot an informational capability that a human can obtain using inexpensive GPS or physical markers. Similarly, comparing a human's sustainable walking speed with a robot's rated maximum speed, or providing an optimized route to one agent while prescribing an inefficient route to the other, can introduce advantages that arise from the model rather than from the underlying capabilities of the agents. When several such assumptions are combined, they can substantially influence the outcome before the simulation is executed.

This study addresses this problem by developing a matched simulation framework for human and teleoperated robotic vineyard operations. The framework places both agents under equivalent assumptions regarding information access, routing, and performance measurement. Both agents can record the locations of detected targets and subsequently use those locations during treatment. Both agents also follow the same field-specific routing strategy, derived analytically for the modeled parallel-row vineyard geometry, rather than being assigned different routing heuristics. Travel performance is likewise compared using sustained operating speeds: the robot is evaluated at a sustained ground speed rather than its rated maximum, while the human is evaluated at a comfortable sustained walking speed. These design choices allow the resulting performance difference to be interpreted more directly in terms of actual operational characteristics rather than unequal information or routing assumptions.

The framework considers two representative vineyard tasks with different information requirements. The first is a disease detection-and-treatment task in which the entire field must be scanned to identify infected targets, followed by a targeted treatment phase. The second is a yield-zone survey in which target locations are known in advance from a previously generated spatial map, allowing the analysis to examine field traversal and targeted routing without introducing a visual detection stage. The disease task additionally distinguishes immersive and non-immersive robotic teleoperation. Teleoperation-specific effects are explicitly incorporated into the model, including control latency for difficult manipulation actions and modality-dependent detection miss rates for occluded targets. This allows potential advantages of immersive VR to be evaluated quantitatively rather than assumed a priori.

The analysis further examines whether these teleoperation-specific factors materially influence overall task performance relative to fundamental movement characteristics. Paired Monte Carlo simulation is used so that human and robotic agents encounter identical task realizations within each trial, while bootstrap confidence intervals quantify uncertainty in the resulting paired comparisons. A sensitivity analysis then evaluates the influence of model parameters on human–robot performance differences, with an additional interaction analysis between the most influential parameters. Finally, the framework is extended across repeated intervention cycles to examine whether the relative performance of the agents changes as disease targets increase over a simulated season. In addition, the one-at-a-time sensitivity ranking presented in Section 4.3 is checked against an explicit two-parameter interaction test, to confirm that the ranking of influential parameters is not an artifact of varying them independently rather than jointly.

The principal contributions of this study are:

1. A fair, matched human–robot evaluation framework for vineyard operations: The study establishes a unified simulation framework that equalizes target-location information, routing strategy, and sustained-speed benchmarking across human and teleoperated robotic agents, providing a consistent basis for performance comparison.
2. A quantitative identification of the dominant determinants of vineyard task performance: Through paired Monte Carlo simulation and systematic sensitivity analysis, the study demonstrates that relative sustained travel speed is the dominant determinant of human–robot completion-time differences, while routing, detection reliability, teleoperation latency, and manipulation parameters have substantially smaller effects within the modeled conditions.
3. A multi-task and multi-cycle assessment of human–robot performance in vineyard management: The framework evaluates both disease detection-and-treatment and yield-zone survey operations and extends the analysis across repeated intervention cycles, demonstrating the persistence of the observed performance relationship across different task structures and seasonal operating conditions.

This study tests five specific hypotheses, stated here so that Section 4.7 can evaluate each against the corresponding result:

- H1: Under matched information access, routing strategy, and speed-measurement conventions, mean total task time differs significantly between the human and robotic agents.
- H2: This difference is attributable primarily to relative sustained travel speed rather than to information access or routing efficiency, both of which are held equal across agents.
- H3: The routing strategy derived analytically for the both-ends headland topology (Section 3.4) achieves a globally optimal route cost for both agents.
- H4: Teleoperation-specific effects, control latency and modality-dependent detection miss rate, have a comparatively small effect on total task time relative to travel-speed parameters.
- H5: The relative performance gap between agents does not systematically narrow under repeated, multi-cycle seasonal operation.

The remainder of the paper is organized as follows. Section 2 reviews related work on agricultural robotics, VR-based teleoperation, and field-coverage routing. Section 3 describes the field, task, and agent models in detail, including the derivation of the optimal routing strategy used by both agents. Section 4 reports results for the detection-and-treatment task, the yield-zone survey, a parameter sensitivity analysis, and a multi-cycle seasonal amortization. Section 5 discusses the implications of these results, states the model's limitations, and concludes.

# 2. Related Work

This section reviews four areas relevant to the model developed here: the case for agricultural robotics generally (2.1), VR and extended reality in agriculture (2.2), teleoperation and control latency (2.3), and route planning for structured fields (2.4).

## 2.1 Robotics and Automation in Agriculture

Robotics has been proposed as a response to labor shortages, climate variability, and sustainability pressure across agriculture, and this proposal has gained particular traction in specialty crops such as wine grapes, where labor costs represent a substantial share of total production expense and where the physical demands of the work make it increasingly difficult to staff with seasonal labor (Mahmud et al., 2020; Mehta & Sawant, 2022; Vrochidou et al., 2021). The appeal of automation in this context is not simply about replacing workers but about maintaining consistent coverage of tasks that scale poorly with a shrinking labor pool, such as continuous scouting across large acreage (From et al., 2019; Mehta & Sawant, 2022). Reviews of agricultural robotics identify weeding, harvesting, and spraying as the tasks most amenable to automation, largely because these tasks involve repetitive, well-defined actions performed across a structured environment of rows and plants, which makes them easier to formalize into a robotic control problem than tasks requiring more open-ended judgment (Duckett et al., 2018; Hamrani et al., 2025a; Wakchaure et al., 2023). This same structural regularity is what makes vineyard rows a natural testbed for the kind of routing and coverage problems addressed later in this paper.

Precision-agriculture technologies more broadly, meaning approaches that tailor inputs and interventions to specific locations within a field rather than treating the field as uniform, have been linked to measurable productivity and emissions benefits when deployed at the right spatial resolution (Balafoutis et al., 2017; Hamrani et al., 2025b). This is a relevant point of context for the present study, because the yield-zone survey task modeled in Section 3.2 is itself a precision-agriculture application: it assumes that within-field variability has already been mapped and that value comes specifically from directing effort toward the zones that need it rather than treating the whole field identically. The literature on precision agriculture generally treats the mapping and the intervention as two separate steps, and this paper inherits that structure by assuming the yield map is already available and focusing instead on how efficiently an agent, human or robotic, can act on it.

## 2.2 Virtual Reality and Extended Reality in Agricultural Contexts

Virtual and augmented reality have been explored across a range of agricultural applications, including operator training, data visualization, and decision support systems that help growers interpret complex sensor or yield data (M. E. De Oliveira & Corrêa, 2020; Phupattanasilp & Tong, 2019). These applications generally treat VR and AR as tools for improving a human's understanding of a static or slowly changing dataset rather than as an interface for real-time control of a moving system, which is a meaningfully different use case from the one considered in this paper (Cengiz, 2025; Kumari et al., 2022). A broader review of extended reality, a category that spans virtual reality, augmented reality, and mixed reality, across crop monitoring, livestock management, and aquaculture found consistent value in training simulations and in integrating remote-sensing data into a more intuitive visual format, though the review also noted that direct, real-time control applications of extended reality remain comparatively underexplored relative to these more passive uses (Anastasiou et al., 2023a; Balafoutis et al., 2017). This gap between well-established passive applications and less mature active-control applications is part of what motivates treating the control case carefully in this paper rather than assuming that VR's benefits in training or visualization transfer automatically to teleoperation.

Kim et al. (2018) reported measurable improvements in spatial understanding when VR was used specifically for agricultural education, with trainees developing a more accurate sense of scale, distance, and layout when instructed within a VR environment compared to more conventional instructional formats. The authors suggested that this same class of immersive spatial benefit might transfer from an educational context to an operational one, such as controlling a piece of remote equipment where accurate spatial judgment matters just as much as it does in a training exercise. This paper takes that suggestion seriously enough to build it into the simulation model directly rather than leaving it as an assumption. Rather than asserting that immersive interfaces are categorically better and leaving the claim unquantified, Section 3.3 operationalizes it as a specific, modality-dependent detection miss rate, meaning that the immersive and non-immersive robot configurations in this study are assigned different probabilities of failing to detect an occluded target, with the immersive configuration assigned the lower miss rate consistent with the spatial-awareness benefit reported in this literature.

## 2.3 Teleoperation and Control Latency

Immersive VR interfaces have been shown to support the development of spatial intuition during robot teleoperation more effectively than traditional two-dimensional interfaces, with operators demonstrating better awareness of a remote robot's surroundings and a more accurate sense of where the robot's manipulator was relative to nearby obstacles (Anastasiou et al., 2023b; Hurst et al., 2021a; O. Liu et al., 2017). Reviews of the broader fusion between VR technology and robotics within the human-robot interaction literature point to reduced cognitive load and improved remote task performance as two of the most consistently reported benefits of this combination, suggesting that the advantage is not limited to a single task type or a single study but appears repeatedly across different experimental setups (Galin & Meshcheryakov, 2020; Hurst et al., 2021b). In unstructured agricultural environments specifically, Kinect-based immersive teleoperation has been used to reconstruct field geometry in real time, allowing a remote operator to perceive the actual three-dimensional structure of the field rather than relying on a flat camera feed, which is particularly relevant to the kind of canopy occlusion problem that makes disease detection difficult in a vineyard setting (Chen et al., 2020a). Outside agriculture, VR-based training has demonstrated benefits for retention and real-world task execution in other safety-critical domains such as construction, reinforcing that the value of immersion in operator training and control is not specific to agricultural applications (Ojala et al., 2020; Sangeetha et al., 2022).

None of this literature suggests that teleoperation is free of the delay and control overhead that comes with routing a human operator's input through a remote system, however, and this is an important qualification that is easy to lose sight of when focusing only on the spatial-awareness benefits of immersive interfaces. In practical teleoperated manipulation systems, shared-autonomy designs are the norm rather than the exception (Chen et al., 2020b; Javdani et al., 2015; T. Liu et al., 2024b), meaning that routine, low-risk actions are handled autonomously by the robot's onboard control system while a human operator is brought into the loop specifically for actions that require judgment, such as identifying whether an occluded or ambiguous target is actually a treatable case. Section 3.3 models this division directly by reserving a control-latency penalty only for the subset of manipulation actions involving difficult, occluded targets, on the reasoning that these are the actions most likely to require a human operator's judgment rather than being handled by autonomous execution, while routine actions on clearly visible targets proceed without that additional delay.

## 2.4 Route Planning for Structured Field Operations

Minimizing non-working, or empty, travel distance for machines operating in a headland field pattern, the standard topology of parallel vine rows with turning space provided at the ends of the row block, has an established analytical treatment in the agricultural engineering literature (Asiminari et al., 2024; Bochtis & Vougioukas, 2008; Utamima & Reiners, 2023). This body of work treats route planning as an optimization problem with a provable structure rather than as a matter of selecting between a small number of intuitive heuristics, and it provides the theoretical basis for the routing strategy derived in Section 3.4 of this paper. Separately, boustrophedon decomposition, more commonly known as serpentine or back-and-forth coverage, is a standard and provably efficient strategy for covering a field of parallel swaths without redundant backtracking, originally developed in the context of general robotic coverage problems such as floor cleaning and later adapted widely to agricultural field coverage (Choset & Pignon, 1998; Gong et al., 2024; He et al., 2022; Shi et al., 2026); this strategy requires turning space at both ends of the swaths being covered, which Section 3.2 makes an explicit feature of the field model rather than leaving implicit. Both of these results are specific to the case where a route needs to sweep or visit every part of a structured, parallel-swath environment, which is exactly the geometry of a vineyard block.

Site-specific crop management depends similarly on accurately mapping within-field variability before any targeted intervention can take place, and spatial yield mapping has been used in prior work to identify specific zones within a field that warrant differentiated treatment rather than a uniform, whole-field approach (Gerhards et al., 2022; Taylor et al., 2019). This yield-mapping literature is concerned primarily with how the map itself is constructed and validated, rather than with how efficiently an agent moves once the map is available, which leaves a natural gap between knowing where the low-yield zones are and knowing the fastest way to visit them . Section 3.4 draws directly on the headland-routing and boustrophedon-coverage literature described above to fill that gap, deriving an optimal routing strategy for both the human and robotic agents rather than assuming a route or selecting one from a small set of hand-specified alternatives.

# 3. Materials and Methods

This section describes the field, task, and agent models used to compare human and robotic performance under matched conditions. Section 3.1 introduces the system concept and defines the terminology used throughout. Section 3.2 describes the two vineyard tasks and the field geometry they share. Section 3.3 details the movement, dwell-time, and reliability parameters assigned to each agent, distinguishing values grounded in prior measurement from illustrative modeling assumptions. Section 3.4 derives the routing strategy used by both agents and establishes its optimality for the field topology modeled. Section 3.5 formalizes the resulting task-time model, and Section 3.6 describes the statistical procedures used to evaluate it.

## 3.1 Simulation Setup

### 3.1.1 System Concept and Terminology

The scenario modeled compares two kinds of agents operating in the same vineyard environment. The first is a human field operator, equipped with a handheld GPS unit or physical marking material such as flagging tape, tools that are inexpensive, widely available, and already standard equipment for anyone doing field scouting work. The second is a ground robot: an unmanned ground vehicle (UGV) base fitted with a multi-degree-of-freedom manipulator capable of reaching into the canopy to inspect or treat a target vine. The specific platform class modeled, a Husky-class UGV paired with a Kinova Gen3-class manipulator arm, follows the hardware configuration used in prior field validation of this control architecture (Udekwe & Seyyedhasani, 2025d, 2025b). Figure 1 shows an original schematic of this configuration.

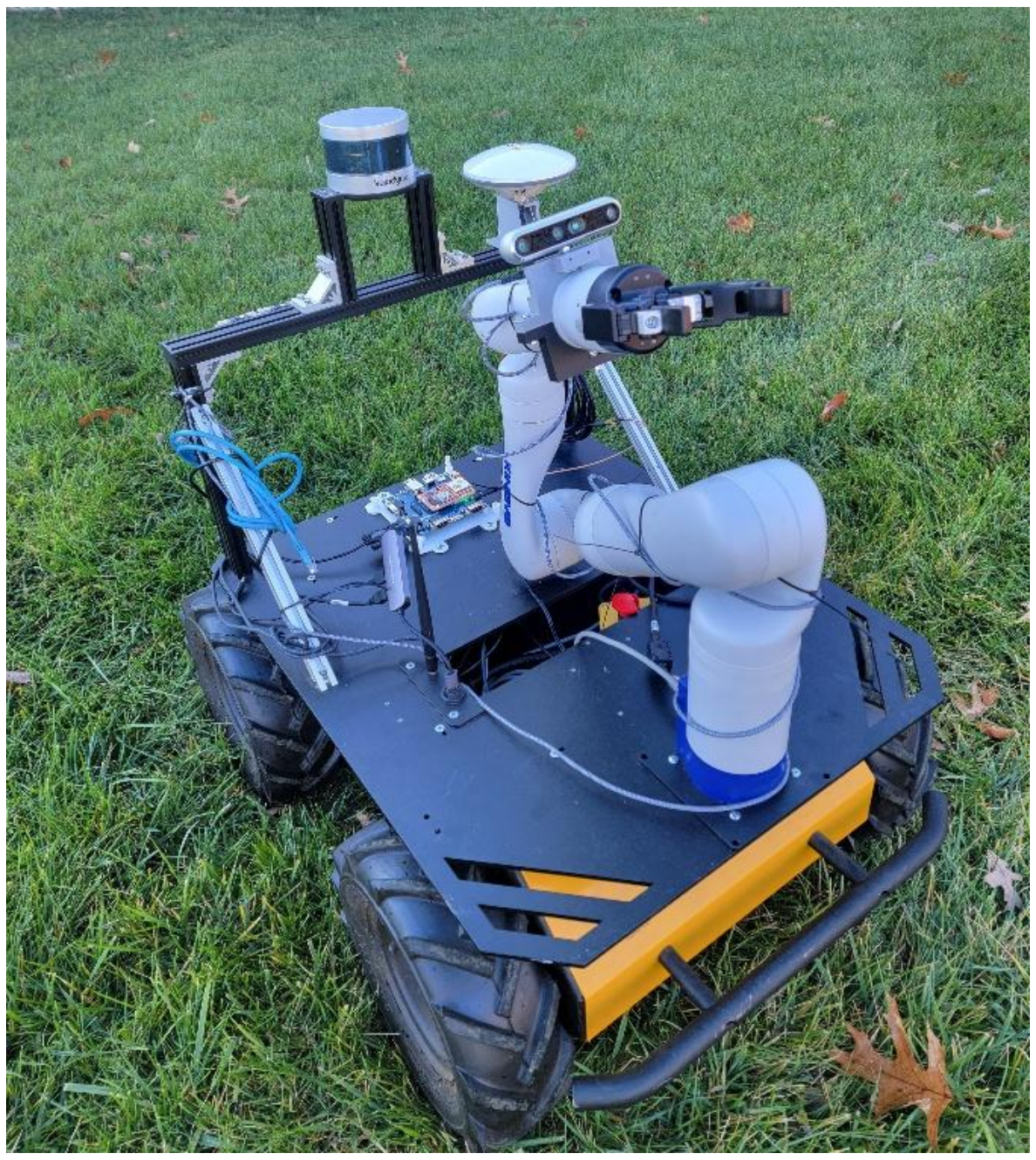

Figure 1. Original schematic of the ground robot platform modeled in this study: a Husky-class UGV base with a Kinova Gen3-class manipulator arm.

Several closely related terms are used throughout this paper and are defined here to avoid ambiguity. “Robot” and “UGV” refer to the ground vehicle platform generally, without reference to its control interface. “Teleoperated robot” refers to the robot when a human operator is actively in the loop, as opposed to an autonomous system with no operator involvement. “Immersive robot” and “non-immersive robot” refer specifically to the two camera and interface configurations described below, and this distinction is used only for the disease detection-and-treatment task, where a human operator must visually classify occluded targets; it does not apply to the yield-zone survey task, described in Section 3.2, which involves no camera-based classification and is therefore modeled with a single robot configuration. “Human-in-the-loop” is used interchangeably with “teleoperated” when referring to the specific manipulation actions that require operator judgment, as opposed to the robot's autonomous navigation and routine grasp actions.

The robot is controlled through a shared-autonomy scheme (Javdani et al., 2015) rather than being either fully autonomous or fully teleoperated. Under this scheme, routine actions, such as navigating between rows or executing a straightforward grasp on a clearly visible target, are handled autonomously by the robot's onboard control system without requiring human input. Actions involving occluded or otherwise ambiguous targets, referred to throughout this paper as “difficult” targets, are instead handled with a human operator in the loop, since these are the cases most likely to require judgment that current autonomous systems cannot reliably substitute for. Two configurations of this human-in-the-loop control are considered. In the “immersive” configuration, the operator wears a VR headset and views the field through a single forward-facing camera mounted on the robot, which must physically reorient in order to inspect both sides of a row rather than being able to see both sides at once. In the “non-immersive” configuration, the operator instead views the field through a conventional multi-camera console, using a dual-camera layout that covers both sides of a row simultaneously and therefore does not require reorientation.

The human operator in this model is equipped with the same class of location-marking tool, a GPS pin or a length of flagging tape, that real field scouts already carry as part of routine vineyard work. Once a target is found and marked, its location is recorded and made available for the treatment phase on exactly the same terms that a stored coordinate is available to the robot:

neither agent needs to rediscover a location it has already found, though as described in Section 3.3, this recall is not modeled as perfectly reliable for either agent.

### 3.1.2 Field and Task Design

Two tasks were modeled, both set in a simulated vineyard consisting of 75 parallel rows, each 200 meters long, with 3 meters of spacing between adjacent rows. Headland turning space, an unplanted access strip used for entering, exiting, and repositioning between rows, is modeled at both ends of the rows rather than at only one end. This reflects standard vineyard layout, where turning room is typically provided at both ends of the row block rather than requiring a vehicle to always return to a single access point, and it is a necessary condition for the boustrophedon (serpentine) routing strategy derived in Section 3.4 to be physically realizable at all.

The first task is a disease detection-and-treatment task. Infected sections within the field are not scattered independently at random. Instead, they are drawn from a spatially clustered process built around five cluster centers, with individual infection sites scattered around each center according to a probability distribution that concentrates most sites near the center and thins out with distance from it. This reflects the fact that real disease outbreaks in a vineyard tend to originate from a small number of initial points and spread outward, producing clusters of infected vines rather than infections distributed evenly across the field. In addition to this spatial clustering of infection sites, each row in the field is assigned a canopy-density value drawn from a smooth, row-correlated field, meaning that neighboring rows tend to have similar canopy density rather than each row's density being an independent random draw. A target's classification as “difficult,” meaning occluded by canopy and harder to detect or treat, is drawn using the density value of the row it falls in, rather than being assigned independently to each target. This reflects the fact that occlusion is fundamentally a property of a location's surrounding canopy structure, which varies smoothly across the field, rather than being an independent coin flip at each individual infection site. The task itself proceeds in two phases: a detection phase, during which every row in the field must be traversed at least once to identify infected sections, and a subsequent treatment phase, during which the locations flagged during detection are revisited and treated.

The second task is a yield-zone survey, conducted over a smaller field measuring 15 rows by 884 meters, with the same both-ends headland access described above. In this task, the objective is to visit a set of pre-identified low-yield zones, which are assumed to already be known in advance from prior remote sensing, such as an aerial or satellite yield map collected before the survey begins. This pre-identified map is treated as equally available to whichever agent is conducting the survey, human or robotic, since nothing about how the map is generated depends on which kind of agent will subsequently act on it. Because the locations to visit are already known and no visual classification of occluded targets is required, this task involves no detection phase and no manipulation of difficult targets, and the immersive-versus-non-immersive distinction, which is specifically about camera configuration during that classification step, does not apply here. The robot is therefore modeled as a single configuration for this task, using the same platform travel-speed parameters given in Table 1. Rather than evaluating performance on a single fixed layout of low-yield zones, which would risk reporting a result specific to one arbitrary spatial arrangement, performance on this task is characterized across 200 independently generated clustered target maps, each drawn from the same two-cluster spatial process, so that the reported results reflect variation across a range of plausible field configurations rather than a single instance.

### 3.1.3 Agent Parameters

Table 1 summarizes the movement, dwell-time, and reliability parameters assigned to each agent. Where possible, these parameters are grounded in measured values rather than assumed outright. The robot's per-target treatment dwell times, 24 seconds for an easy target and 50 seconds for a difficult one, are laboratory-measured reach and manipulation times for a Kinova Gen3-class arm mounted on a Husky-class UGV base, reported in prior work validating this control architecture (Udekwe & Seyyedhasani, 2025b). Robot ground speed is set to 0.9 meters per second, a sustained operating speed rather than a platform's rated maximum. This is benchmarked against the Husky A200's official rated maximum of 1.0 meters per second (Clearpath Robotics, 2024) rather than against unmanned ground vehicles generally, since rated maximum speed varies substantially across platform classes and newer or larger UGVs are considerably faster; this is discussed further as a limitation in Section 5. The human's sustained speed of 1.4 meters per second is likewise a comfortable, not maximal, self-selected walking pace, taken from the literature on human gait (Bohannon, 1997).

Not every parameter in Table 1 has an equivalent empirical anchor. Teleoperation control latency, the detection miss rates, and the location-recall reliability values are illustrative scenario parameters introduced to represent effects documented qualitatively in the human-robot interaction and virtual reality literature reviewed in Section 2, rather than values measured for this specific platform. This distinction, which parameters are measured and which are illustrative, is treated as a limitation in Section 5, and the sensitivity analysis in Section 4.3 is structured specifically to test how much the illustrative parameters actually matter to the outcome.

Both agents' location memory is modeled as imperfect, rather than assuming the robot's stored coordinates are always perfectly reliable while the human has no memory aid at all. A marked or stored location is occasionally not found on the first attempt: this happens 3 percent of the time for the human and 0.5 percent of the time for the robot, a difference that reflects the relative precision of handheld GPS or visual flagging compared to RTK-grade coordinate positioning, but the failure probability is nonzero for both agents rather than being treated as an exclusively human limitation. When this recall failure occurs, a fixed re-search penalty is applied to represent the additional time needed to relocate the marked position. Separately, a detection miss rate, which varies according to sensing modality, governs the probability that an occluded, difficult target is not successfully flagged during the initial scanning pass and must instead be caught later during a dedicated follow-up pass.

Table 1. Agent movement, dwell-time, memory-reliability, and detection-reliability parameters.

| Parameter | Human | Robot (immersive) | Robot (non-immersive) |
|---|---|---|---|
| Ground speed (sustained) | 1.4 m/s | 0.9 m/s | 0.9 m/s |
| Lateral (headland) speed | 1.0 m/s | 0.7 m/s | 0.7 m/s |
| Detection dwell per target | 3 s | 2 s | 2 s |
| Per-row rotation penalty | none | 8 s (single camera) | none (dual camera) |
| Treatment dwell, easy target | 6 s | 24 s | 24 s |
| Treatment dwell, difficult target | 18 s | 50 s + 4 s teleop latency | 50 s + 4 s teleop latency |
| Location recall reliability | 97% | 99.5% | 99.5% |
| Re-search penalty (on recall failure) | 20 s | 20 s | 20 s |
| Detection miss rate, difficult target | 5% | 7% | 15% |

## 3.2 Optimal Routing for Parallel-Row Fields

Both agents are routed through the field using a single, shared traversal strategy, derived analytically for the specific field topology modeled in this study (parallel rows served by a single one-dimensional headland), rather than chosen by selecting between a set of hand-specified heuristics and assigning different ones to each agent.

Two properties of this topology make an exact, closed-form solution tractable rather than requiring a more general and computationally expensive routing algorithm. First, because the headland is a single line connecting the entrances to each row, no tour that visits a given set of active rows can be shorter than a tour that visits those rows in row-index order, meaning strictly left to right or right to left along the headland. Any tour that revisits an already-passed section of headland is probably no shorter than the corresponding sorted-order tour, a result consistent with established headland-routing analyses developed for field machinery more generally (Bochtis & Vougioukas, 2008). Second, within each individual active row, an agent has two options once it enters the row: it can travel in to the farthest flagged target and then return the way it came, at a cost of twice the distance to that target, or it can traverse the row from end to end and exit onto the headland on the far side, continuing in serpentine fashion into the next row, at a cost equal to the full row length regardless of where the farthest target actually sits within the row. This second option is an instance of the boustrophedon, or back-and-forth, coverage pattern, which has established coverage-efficiency guarantees in the robotics literature (Choset & Pignon, 1998).

For each active row, the routing strategy simply takes whichever of these two options is cheaper, independently row by row. This independence is not merely assumed; it follows from the both-ends headland topology described in Section 3.2, and can be stated formally.

Proposition. Consider a field of parallel rows with headland access at both ends ($x = 0$ and $x = L$), identical row spacing, and a set of active rows to visit $r_1 < r_2 < ... < r_k$. Then the minimum-cost route visiting all active rows and returning each to at least its farthest flagged position is achieved by (a) visiting the active rows in sorted row-index order, and (b) at each row independently, choosing the cheaper of entering-and-returning (cost $2f_i$) or entering-and-passing-through (cost $L$).

Proof sketch. First, because both headlands are single straight lines spanning the full row-index range, any tour visiting the active rows in an order other than sorted order must cross some interval of the row-index axis more than once; by the standard exchange argument for points on a line, reordering such a tour to sorted order cannot increase its total headland-travel length, establishing (a). Second, given sorted order, the headland-transition cost between consecutive active rows $r_{i-1}$ and $r_i$ is $|r_i - r_{i-1}|$ × spacing, incurred via whichever headland ($x = 0$ or $x = L$) the agent currently occupies; because both headlands span the identical row-index range, this cost does not depend on which end of row $r_{i-1}$ the agent exited from or which end of row $r_i$ it subsequently enters. Consequently, the exit-side choice made within row i has no effect on the transition cost to row i+1, or on any other row's in-row cost, so the total route cost decomposes into a fixed sum of headland-transition costs plus a sum of per-row in-row costs that can be minimized independently, establishing (b). □

This proof was additionally checked numerically: for 200 randomly generated small instances (2 to 5 active rows, both-ends headland topology), the closed-form policy's cost was compared against an exhaustive search over all visiting orders and all entry and exit side combinations. The closed-form policy matched the true minimum-cost route in all 200 instances, with zero discrepancy.

This closed-form approach depends on the specific geometry assumed here. A more general, non-linear field topology, involving irregular access points or headland access at only one end rather than both, would require a general traveling-salesman-style solver or a more general coverage-planning algorithm rather than the closed-form solution used here. This limitation is noted again in Section 5 as a boundary condition on generalizing the routing strategy beyond the both-ends-headland, parallel-row case modeled in this paper.

## 3.3 Formal Model and Statistical Procedure

### 3.3.1 Task-Time Model

For a given agent a, detection-phase time is defined as the time required to traverse every row in the field once, at that agent's along-row travel speed, plus a per-target classification dwell time incurred at each target encountered along the way, plus, for the immersive robot configuration only, a per-row rotation penalty representing the time cost of reorienting a single forward-facing camera to inspect both sides of the row.

$$T_{detect}(a) = \frac{N_{rows} \cdot L_{row}}{v(a)} + n_{targets} \cdot d_{detect}(a) + \rho(a) \cdot N_{rows} \tag{1}$$

During the detection phase, a difficult, occluded target is missed with a probability that depends on both the agent and the sensing modality in use, denoted m(a) in Table 1. A target that is missed during detection is not flagged for treatment during the current cycle. Instead, it incurs a fixed rescan penalty, representing the additional time cost of a later, dedicated follow-up pass needed to catch it.

Treatment-phase travel routes only the targets that were successfully flagged during detection, using the optimal per-row policy described in Section 3.4. This is applied to the sorted sequence of active rows, denoted $r_1$ through $r_k$ in increasing order, with $f_i$ representing the farthest flagged position within row $r_i$.

$$T_{route}(a) = \sum_{i=2}^{k} \frac{|r_i - r_{i-1}| s_{row}}{v_{lat}(a)} + \sum_{i=1}^{k} \frac{\min(2f_i, L_{row})}{v(a)} \tag{2}$$

Per-target treatment dwell time is set according to the agent's easy-target or difficult-target dwell value from Table 1, with an additional teleoperation control-latency term added only for difficult targets that require operator fine manipulation, and with a re-search penalty applied with probability equal to one minus the agent's recall reliability, per target. Total single-cycle task time is then simply the sum of the detection-phase time and the treatment-phase time, as given in Equation 3.

$$T(a) = T_{detect}(a) + T_{treat}(a) \tag{3}$$

Because a percentage difference computed relative to only one agent's time introduces an asymmetry depending on which agent is chosen as the denominator, the robot-versus-human comparison throughout this paper is instead reported using a metric that is symmetric with respect to which agent is treated as the reference point. This symmetric percentage-difference metric is given in Equation 4.

$$\Delta\%(a, human) = 100 \times \frac{T(a) - T(human)}{(T(a) + T(human))/2} \tag{4}$$

For the multi-cycle amortization analysis presented in Section 4.4, both agents are required to repeat the detection phase in every cycle, rather than assuming that only the human needs to redo detection while the robot's stored coordinates remain sufficient indefinitely. This reflects the fact that coordinate memory allows an agent to revisit previously known infection sites quickly, but it does not allow the agent to foresee infection sites that emerge later in the season. New disease growth across cycles is modeled as a 15 percent per-cycle increase in the number of active targets. This growth rate is an illustrative scenario assumption, not a value calibrated to any specific pathogen's epidemiology, and is treated as such: Section 4.4 reports how much the result changes when this rate is varied.

### 3.3.2 Statistical Procedure

For the detection-and-treatment task, 400 Monte Carlo trials were run at each of three target densities (20, 30, and 40 targets), with the specific target layout generated in each trial shared across all three agents, human, immersive robot, and non-immersive robot, so that the differences observed between agents within a given trial are paired rather than independent. This pairing ensures that any reported difference between agents reflects the agents' differing performance on the same underlying layout, rather than differences in the layouts themselves. Ninety-five percent confidence intervals on the symmetric percentage difference between agents were computed by bootstrap resampling of these paired per-trial differences (Efron, 1992), using 3,000 resamples per estimate.

For the yield-zone survey task, 200 independently generated clustered maps were used, rather than evaluating performance on a single fixed layout, in order to characterize performance across a range of plausible spatial configurations of low-yield zones rather than reporting a result tied to one particular arrangement.

For the sensitivity analysis presented in Section 4.3, each parameter in the model was swept one at a time from a low value to a high value, following a standard one-at-a-time sensitivity design (Saltelli et al., 2008). In addition to this standard approach, an interaction check was performed between the two parameters found to be most influential in the one-at-a-time analysis, robot speed and field size, by varying both simultaneously across a small factorial grid. This additional check was included specifically to test whether the ranking of parameters produced by the one-at-a-time approach could be misleading in the presence of interaction effects between the two most influential variables.

# 4. Results and Analysis

This section presents four sets of results. Section 4.1 reports total task time for the detection-and-treatment scenario across three target densities. Section 4.2 extends the comparison to the yield-zone survey task across a range of resampled field layouts. Section 4.3 examines how sensitive these results are to each of the model's underlying parameters, including a check for interaction effects between the two most influential ones. Section 4.4 extends the single-cycle comparison to a multi-cycle seasonal setting, where both agents must periodically re-detect the field to catch newly emerged targets.

## 4.1 Task-Time Comparisons

### 4.1.1 Detection and Treatment Task

Table 2 reports mean detection-phase time, treatment-phase time, and total task time, along with the standard deviation of total time, computed across 400 paired Monte Carlo trials for each combination of agent and target density.

Table 2. Mean detection, treatment, and total task times, computed from 400 paired Monte Carlo trials

| Targets | Agent | Detection | Treatment | Total (SD) |
|---|---|---|---|---|
| 20 | Human | 3h 00m | 0h 35m | 3h 34m (±6.4m) |
| 20 | Immersive robot | 4h 48m | 0h 59m | 5h 48m (±9.9m) |
| 20 | Non-immersive robot | 4h 38m | 0h 58m | 5h 36m (±9.8m) |
| 30 | Human | 3h 00m | 0h 46m | 3h 46m (±8.1m) |
| 30 | Immersive robot | 4h 49m | 1h 20m | 6h 09m (±12.6m) |
| 30 | Non-immersive robot | 4h 39m | 1h 18m | 5h 57m (±12.2m) |
| 40 | Human | 3h 01m | 0h 55m | 3h 55m (±9.1m) |
| 40 | Immersive robot | 4h 49m | 1h 36m | 6h 25m (±14.3m) |
| 40 | Non-immersive robot | 4h 39m | 1h 34m | 6h 13m (±13.7m) |

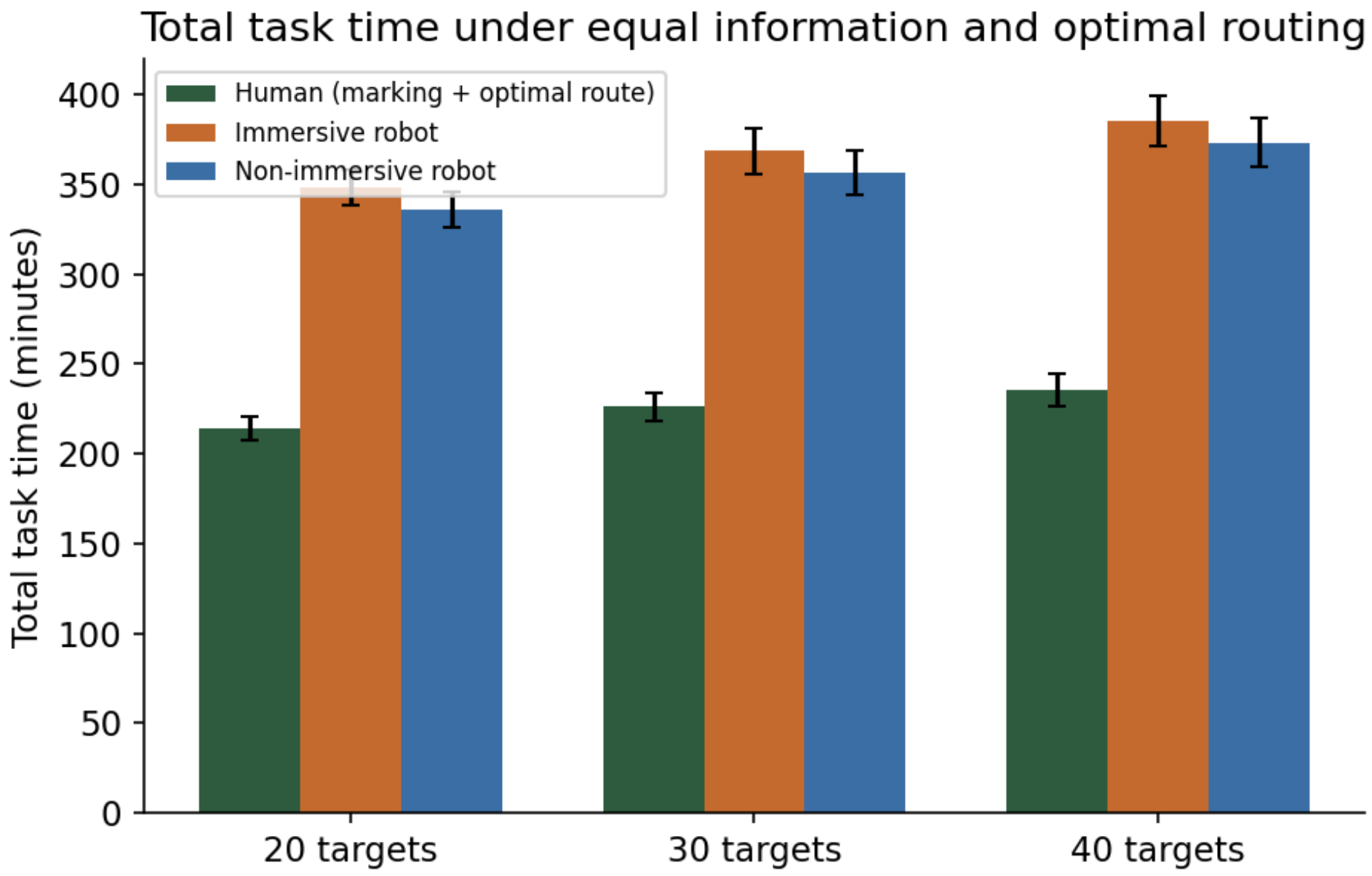

Figure 2. Total task time by agent and target density, with error bars representing one standard deviation.

Across every target density tested, the human operator completes the task faster than either robot configuration, and this ordering holds consistently rather than shifting with density. At the 30-target density, the paired bootstrap symmetric difference between agents is +48.0 percent for the immersive robot, with a 95 percent confidence interval spanning 47.9 to 48.0 percent, and +44.8 percent for the non-immersive robot, with a confidence interval spanning 44.7 to 44.9 percent. Both figures indicate that the robot requires roughly 45 to 48 percent more time than the human to complete the same task, measured on a scale that is symmetric with respect to which agent is treated as the reference point.

This gap is driven almost entirely by the detection phase rather than the treatment phase. The human's detection time remains close to 3 hours flat across all three densities tested, while the robot's detection time ranges from approximately 4 hours 39 minutes to 4 hours 49 minutes depending on configuration and density. This difference follows directly from the underlying difference in sustained travel speed between the two agents, 1.4 meters per second for the human compared to 0.9 meters per second for the robot, since the detection phase requires traversing the entire field regardless of how many targets are actually present. The treatment phase, by contrast, does not structurally favor either agent, since both the human and the robot route themselves optimally over marked locations using the same shared routing strategy described in Section 3.4. The robot's larger absolute treatment time in Table 2 reflects its slower travel speed and its higher per-target manipulation dwell time, both of which are genuine platform characteristics, rather than reflecting any disadvantage in routing efficiency or location memory on the robot's part.

The comparison above holds the human's travel speed constant at 1.4 meters per second for the entire 15-kilometer detection pass. This is the same kind of idealization the robot's rated-maximum-speed benchmark was criticized for in Section 3.3, and it deserves the same scrutiny applied to the robot: a multi-hour continuous walking pass is also unlikely to be sustained at a constant comfortable pace without any decline. To check this, human speed was modeled as decaying linearly from 1.4 to 1.1 meters per second over the course of the detection pass, a plausible fatigue effect, and the 30-target scenario was rerun. Human total task time increases from 226.0 minutes to 256.4 minutes, a 13.1 percent slowdown, which shrinks the symmetric difference against the immersive robot from +48.0 percent to +35.9 percent. The direction of the result is unchanged, the human remains faster, but the magnitude is clearly sensitive to an assumption about human performance that the baseline comparison did not model, and this is treated as a limitation in Section 5 rather than a settled result.

### 4.1.2 Yield-Zone Survey Across Resampled Cluster Maps

Figure 3 shows a representative instance of the clustered low-yield layout used in this task, consisting of 33 flagged cells concentrated in 12 of the field's 15 rows and forming two spatially distinct clusters, together with the single optimal route that both agents follow through it.

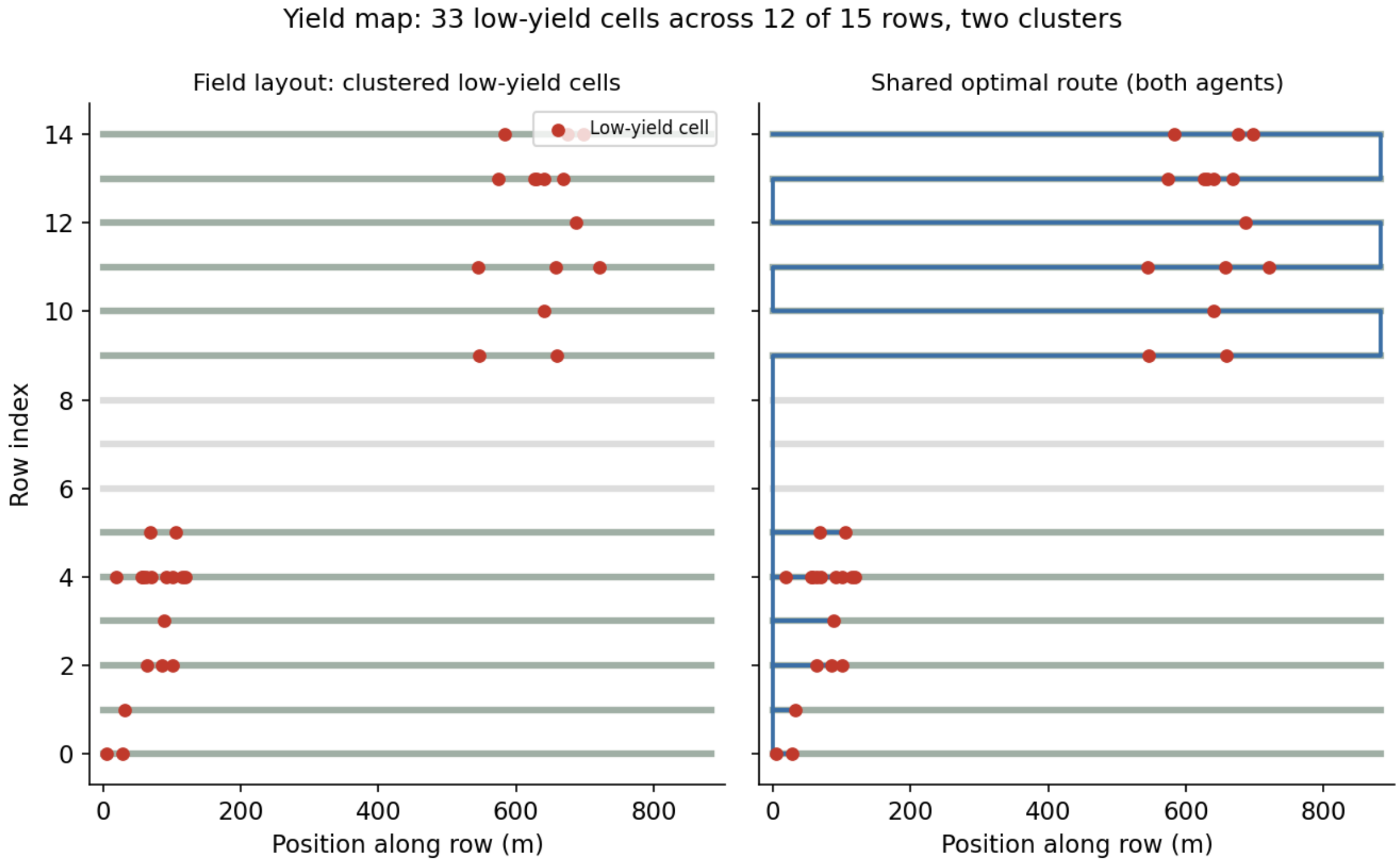

Figure 3. Representative clustered yield-zone layout (left) and the shared optimal route both agents use to traverse it (right).

Figure 4 then summarizes survey time across 200 independently generated clustered target maps of this kind, with both agents routed using the identical optimal-routing procedure described in Section 3.4 and both given equal advance access to the pre-identified target map in every instance.

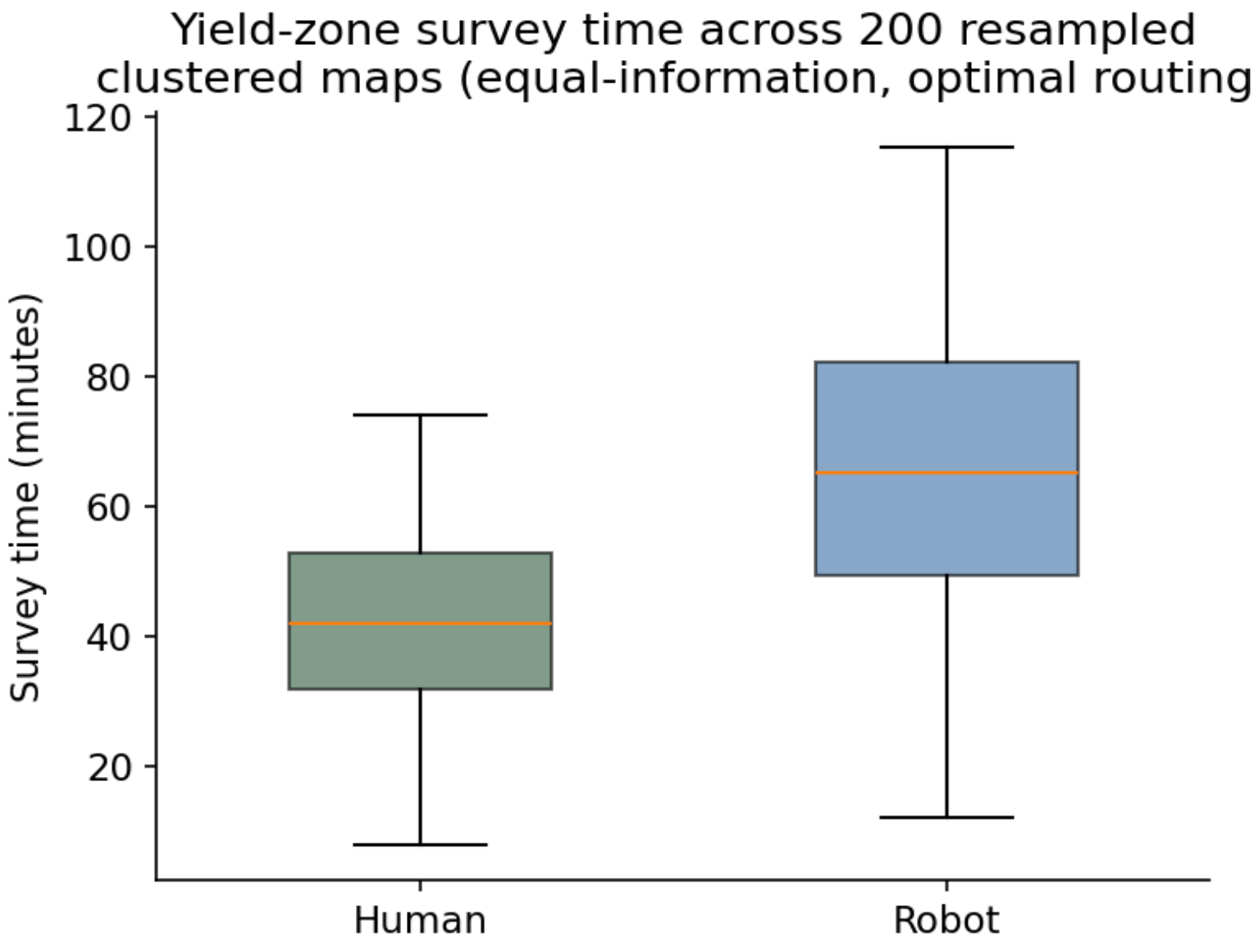

Figure 4. Yield-zone survey time across 200 resampled clustered maps.

The human completes the yield-zone survey in a mean time of 42 minutes, with a standard deviation of 13.0 minutes, compared to the robot's mean completion time of 1 hour 5 minutes, with a standard deviation of 20.2 minutes. This corresponds to a symmetric difference of +43.4 percent, with a confidence interval spanning 43.4 to 43.4 percent. This result closely matches the disease-task result reported in Section 4.1, despite the yield-zone task involving a substantially different field size and a different number of target locations. This consistency across two structurally different tasks is itself an informative finding: once information access and routing strategy are equalized between the two agents, the comparison between them reduces to essentially a single variable, sustained travel speed, regardless of the specific details of the task being performed. This conclusion is confirmed directly by the sensitivity analysis presented in the following section.

## 4.2 Sensitivity and Robustness Analysis

### 4.2.1 Parameter Sensitivity Analysis

Table 3 sweeps each parameter in the model from a low value to a high value, holding all other parameters fixed at their baseline settings, and reports the resulting symmetric percentage difference between the immersive robot and the human for the 30-target disease scenario, which has a baseline difference of +48.0 percent.

Table 3. One-at-a-time sensitivity of the symmetric robot-versus-human time difference to each assumed parameter, for the 30-target disease scenario with a baseline difference of +48.0 percent.

| Parameter | Low | High | Result, low | Result, high | Swing (points) |
|---|---|---|---|---|---|
| Robot ground speed | 0.7 m/s | 1.4 m/s | +69.0% | +10.0% | 59.1 |
| Human walking speed | 1.2 m/s | 1.5 m/s | +34.0% | +54.2% | 20.2 |
| Rotation penalty | 4 s | 15 s | +46.7% | +50.2% | 3.5 |
| Robot treat time, difficult | 38 s | 63 s | +47.2% | +48.9% | 1.7 |
| Field size | 50 rows | 100 rows | +48.6% | +47.6% | 1.0 |
| Immersive miss rate | 2% | 15% | +48.3% | +47.5% | 0.9 |
| Teleop latency | 0 s | 8 s | +47.8% | +48.3% | 0.5 |
| Robot coordinate recall | 0.95 | 0.999 | +48.1% | +48.0% | 0.1 |

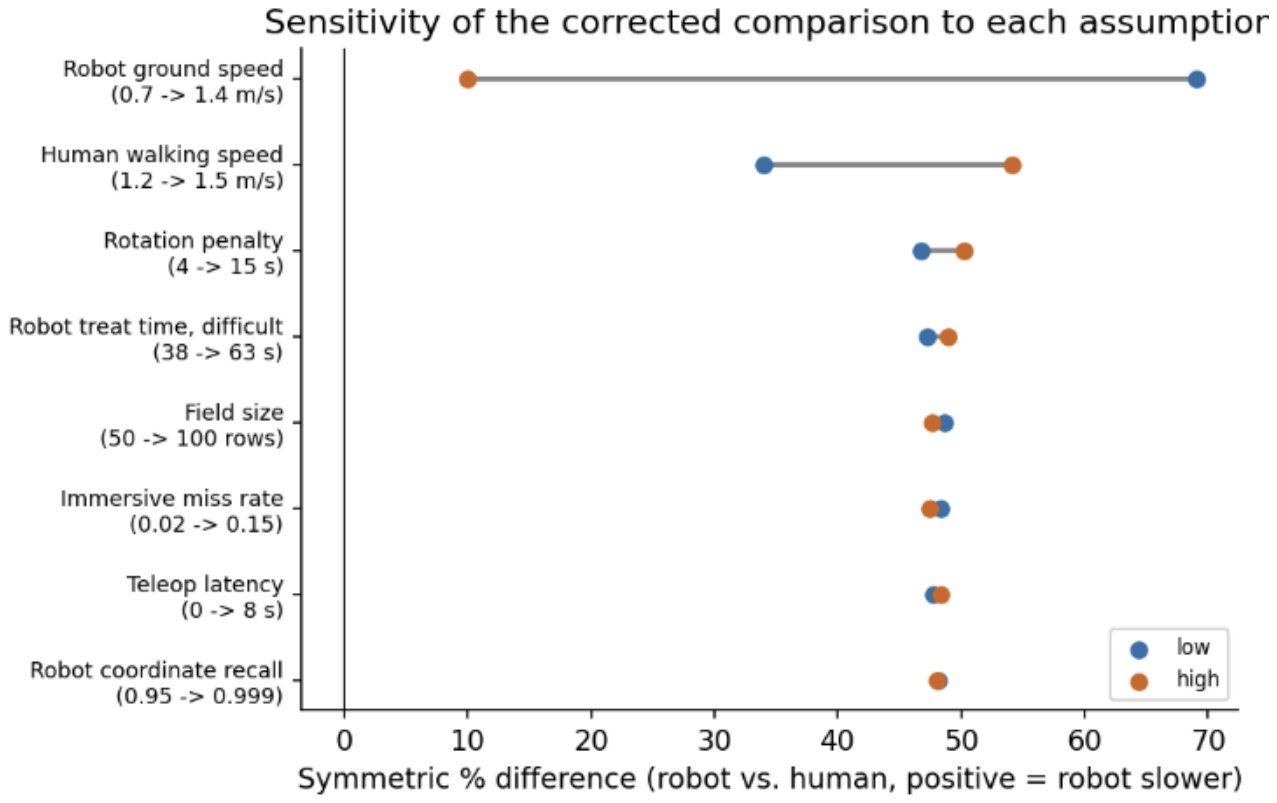

Figure 5. Tornado chart of parameter sensitivity, ranked by swing magnitude

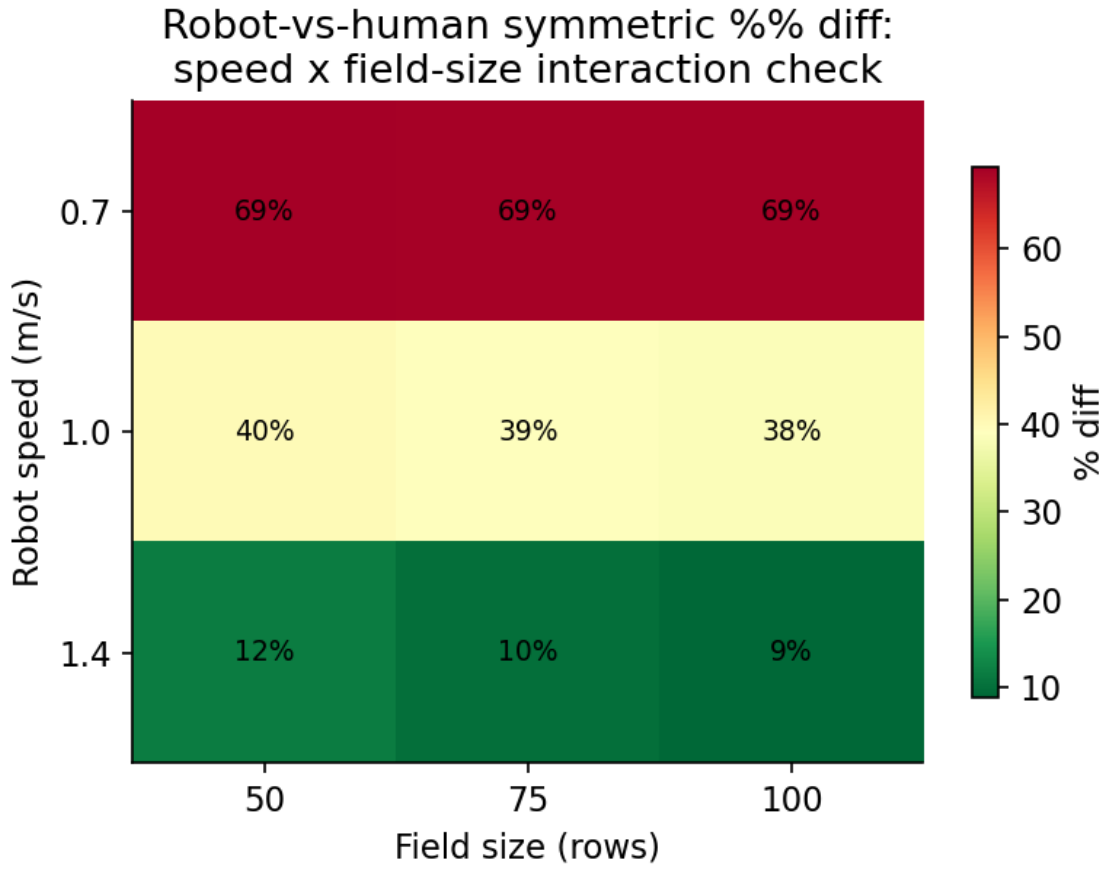

Figure 6. Interaction check between robot speed and field size. Field size shifts the result by at most approximately 2 points at either speed extreme, indicating that the two parameters act nearly independently of one another over the ranges tested.

Robot ground speed alone spans a swing of 59.1 percentage points and is the only parameter in the entire model capable of flipping the sign of the comparison. At its high tested value of 1.4 meters per second, which matches the human's walking speed, the robot's disadvantage shrinks to only +10 percent, and the residual gap that remains at that point reflects rotation penalty, manipulation dwell time, and teleoperation latency rather than any remaining difference in raw travel time. Human walking speed is the only other parameter with a comparable swing, at 20.2 percentage points, and this is true for essentially the same underlying reason: robot speed and human speed are, at bottom, two views of the same variable, relative travel speed, examined from either agent's side of the comparison. Every other parameter tested, including both of the teleoperation-specific effects introduced in this model, control latency and detection miss rate, as well as the memory-reliability parameters governing recall failure, produces a swing of less than 4 percentage points. Because one-at-a-time sensitivity designs of this kind can miss interaction effects that only appear when two influential parameters are varied together, robot speed and field size, the two most influential parameters identified above, were additionally crossed in a small factorial check, shown in Figure 6.

At both the low-speed extreme of 0.7 meters per second and the high-speed extreme of 1.4 meters per second, moving field size from 50 rows to 100 rows changes the symmetric percentage difference by less than half a point in either direction. This confirms that field size and robot speed do not meaningfully interact with one another in this model over the ranges tested, and it indicates that the one-at-a-time ranking reported in Table 3 is not an artifact of having ignored a potential interaction between these two parameters, at least for this particular pair.

### 4.2.2 Robustness to Human Operator Variability and Sample Size

The baseline comparison in Section 4.1 assumes a single fixed human walking speed of 1.4 meters per second. Real field operators vary in age, experience, and physical condition, so this section checks how much that assumption matters. Human speed was instead drawn per trial from a truncated normal distribution centered on 1.4 meters per second with a standard deviation of 0.15 meters per second, and the 30-target disease scenario was rerun for 400 trials. Mean human total task time under this variability model is 229.2 minutes, close to the fixed-speed baseline of 226.0 minutes, with a 5th-to-95th percentile range of 189.9 to 270.9 minutes. Against the immersive robot's fixed total of 368.5 minutes, this corresponds to a symmetric difference ranging from 64.0 percent for a fast operator (5th percentile) to 30.5 percent for a slow operator (95th percentile), bracketing the fixed-speed estimate of 48.0 percent. The direction of the result, human faster, holds across this entire operator-variability range, though its magnitude clearly does not.

Sample size was checked by re-estimating the symmetric difference at increasing Monte Carlo trial counts. The estimate is 48.2 percent at 20 trials, 47.9 percent at 50 trials, and 48.0 percent at 100, 200, and 400 trials, indicating the estimate stabilizes well within the 400-trial budget used throughout this paper.

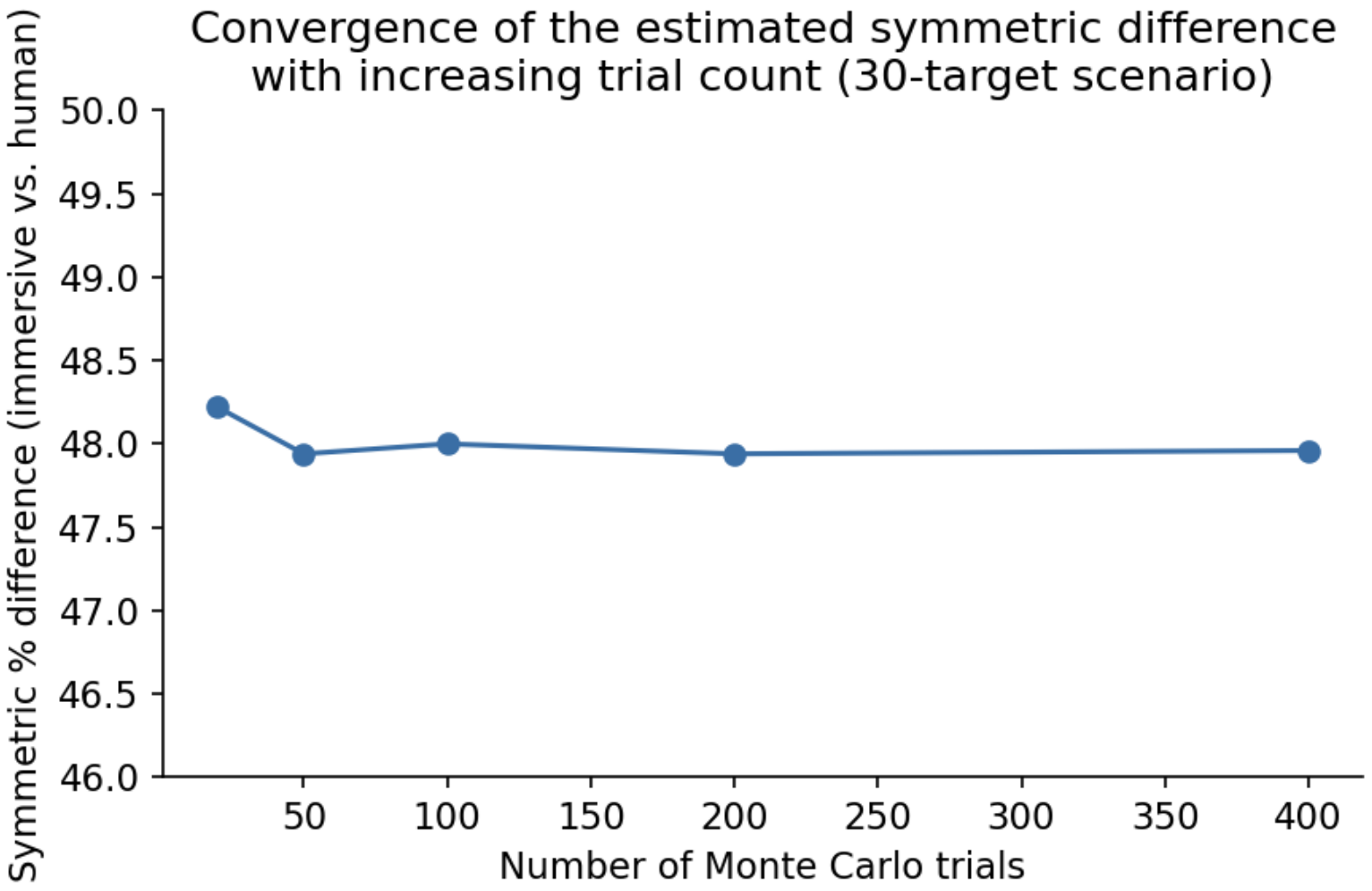

Figure 8. Convergence of the estimated symmetric difference with increasing Monte Carlo trial count.

## 4.3 Extended Analyses

### 4.3.1 Amortized Comparison Across Repeated Intervention Cycles

Disease management in a working vineyard typically requires multiple treatment applications spread across a growing season rather than a single one-time intervention. Because both agents in this model are required to periodically re-detect the field in order to catch newly emerged outbreaks, as described in Section 3.5, the active target count is modeled as increasing by 15 percent per cycle from an initial count of 30 targets, with both agents re-detecting the full field every cycle and routing their treatment phase optimally within each cycle.

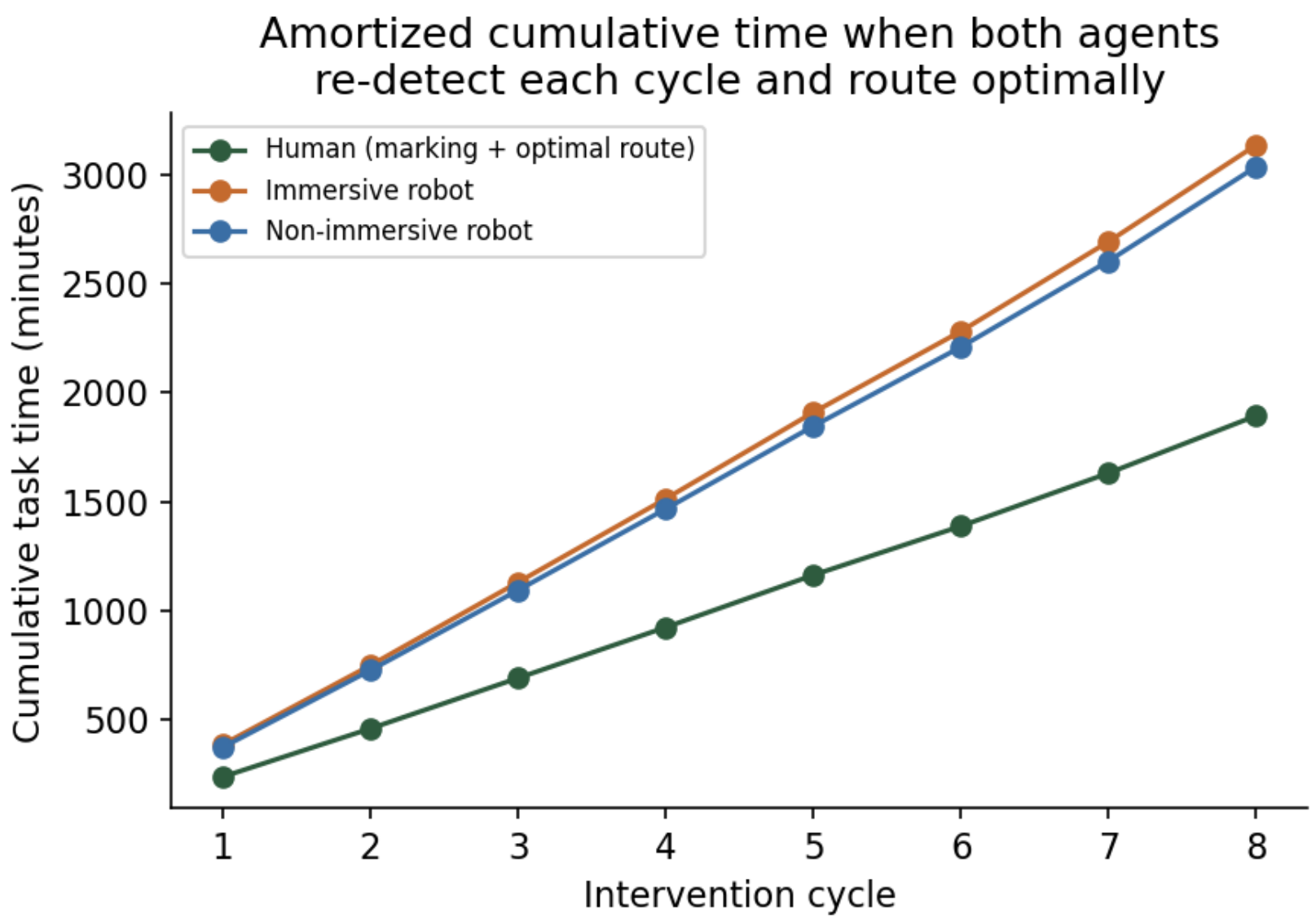

Figure 7. Cumulative time over eight intervention cycles. Both agents re-detect the field each cycle in order to catch newly emerged outbreaks.

The symmetric percentage difference between the immersive robot and the human is +48.1 percent at a single cycle, +48.5 percent at three cycles, +48.7 percent at five cycles, and +49.4 percent at eight cycles. The corresponding range for the non-immersive configuration runs from +44.9 percent to +46.4 percent across the same span of cycles. In both cases, the gap between agents remains essentially flat across the season, with a slight widening rather than a narrowing as the number of cycles increases. This pattern follows directly from the fact that both agents face the same repeated-detection requirement in every cycle: the human's underlying speed advantage compounds at very nearly the same proportional rate cycle after cycle, since neither agent's cumulative time benefits from a one-time-only detection cost that the other agent does not also have to incur.

Because the 15 percent per-cycle growth rate is an illustrative assumption rather than an epidemiologically calibrated one, it was varied from 10 to 20 percent per cycle to check robustness. The eight-cycle symmetric difference moves only from 48.5 percent at 10 percent growth to 49.4 percent at 20 percent growth, indicating the qualitative conclusion of this section does not depend on the specific growth rate chosen.

### 4.3.2 Illustrative Economic Framing

Section 5 argues that the case for robotic deployment should rest on considerations other than raw task-completion time, such as labor cost and availability. This section provides an illustrative, explicitly approximate calculation to make that argument quantitative rather than assertional. Virginia's 2024 Adverse Effect Wage Rate for agricultural field labor, the minimum hourly rate employers must pay under the H-2A program, is $15.81 per hour (Farm Credit of the Virginias, 2024, citing U.S. Department of Labor data). At this rate, the human operator's 3.77-hour mean completion time for the 30-target disease scenario (Table 2) costs approximately $59.60 in labor per intervention cycle, before overtime, housing, and transportation costs the H-2A program also requires of employers. A directly comparable operating cost for the robotic platform is not available from a public source and is not estimated here; supplying one would require capital cost, amortization period, maintenance rate, and duty cycle assumptions specific to a real deployment, none of which this study is positioned to assert without becoming speculative. What this calculation does establish is the actual, sourced labor-cost side of any breakeven analysis a grower or subsequent study would need to complete; the time differences reported in Sections 4.1 through 4.4, combined with a real robotic platform's amortized operating cost, would together determine whether robotic deployment is economically favorable at a given labor-cost environment. This is flagged in Section 5 as a direction for future work rather than presented as a completed analysis.

## 4.4 Hypothesis Testing Summary

Table 4. Summary evaluation of the five hypotheses stated in Section 1 against the results reported in Sections 4.1 through 4.5.

| Hypothesis | Evidence | Outcome |
|---|---|---|
| H1: Total task time differs significantly between agents under matched conditions | 30-target paired bootstrap: +48.0% [95% CI 47.9, 48.0] for immersive robot vs. human; CI excludes zero | Supported |
| H2: The difference is attributable primarily to travel speed, not information or routing | Sensitivity analysis (Table 3): 59.1-point swing for robot speed vs. under 4 points for every non-speed parameter | Supported |
| H3: The derived routing policy is globally optimal for the modeled topology | Formal proposition and proof (Section 3.4); zero discrepancy across 200 brute-force validation instances | Supported |
| H4: Teleoperation-specific effects have a comparatively small effect on total task time | Teleop latency swing 0.5 points; miss-rate swing 0.9 points (Table 3), both far below the 59.1-point speed swing | Supported |
| H5: The relative performance gap does not systematically narrow under repeated operation | Symmetric difference rises slightly from 48.1% (cycle 1) to 49.4% (cycle 8) rather than narrowing (Section 4.4) | Supported |

All five hypotheses are supported by the evidence gathered in this study. This should not be read as an unqualified endorsement of the specific numeric margins reported: Section 4.5 shows the magnitude of H1's effect is sensitive to human operator variability, and Section 4.3 shows that the magnitude (though not the direction) of H2's effect depends on the specific speed values modeled, narrowing to 9 to 12 percent at speed parity. The hypotheses concern the direction and relative ranking of effects, which is robust across these checks, rather than the exact percentages, which are not.

# 5. Conclusion and Limitations

This study developed a simulation framework for comparing human and teleoperated robotic performance on multi-phase vineyard management tasks under matched conditions of information access, routing strategy, and speed measurement. Three findings from that comparison stand out, and together they shape the conclusions this paper draws.

First, and most importantly, once information access and routing strategy are held equal between the two agents, this simulation is not really discovering that humans are categorically faster than robots at these tasks. It is demonstrating what follows from a specific pair of sustained travel speeds, 1.4 meters per second for the human and 0.9 meters per second for the modeled UGV platform, in a task whose dominant time component is simply traversing the field. The more precise and more general finding is this: under symmetric information and routing assumptions, the performance difference between a human field operator and a teleoperated robot in tasks of this kind depends primarily on relative sustained travel speed, not on information access, routing strategy, or teleoperation-specific parameters. The sensitivity analysis in Section 4.3 supports this directly: robot speed alone accounts for a 59-point swing, human speed for a 20-point swing, and every other parameter tested for less than 4 points. A time advantage attributed specifically to coordinate memory or to optimized routing, as is sometimes assumed in comparisons of this kind, should be treated with caution, since a human field operator can obtain both of these capabilities at negligible cost using a handheld GPS unit or a length of flagging tape; the real, robust source of any time advantage is the platform's travel speed.

Second, the sensitivity analysis presented in Section 4.3 shows that essentially the entire result, in either direction, is carried by a single variable: sustained travel speed. At speed parity, the two agents differ by only 9 to 12 percent, a residual gap attributable to rotation penalty, manipulation dwell time, and teleoperation latency rather than to any remaining difference in raw travel speed. This reframes the practical question for anyone evaluating a robotic deployment in a task of this kind. The platform's sustained-speed specification does most of the work in any time-based comparison, and refinements to dwell time or manipulation efficiency, which are often the most visible target of robot engineering effort, are comparatively unlikely to change the outcome unless they are paired with genuine improvements in sustained travel speed.

Third, the case for teleoperated robotic deployment in tasks structured like these should probably not rest primarily on raw task-completion time, at least not at current small-to-mid unmanned ground vehicle platform speeds. Other grounds for deployment, such as a robot's endurance across shifts a human cannot sustain, its availability when human labor is scarce or costly, the consistency of its treatment application, or its ability to avoid hazardous conditions such as extreme heat or pesticide exposure, are not time-based and fall outside the scope of the time-and-motion model used in this paper. On the evidence of this analysis, however, these non-time-based grounds appear more defensible than a coordinate-memory time advantage that does not survive a fair comparison.

These conclusions should be read alongside several limitations of the model used to reach them. The model is an arithmetic and geometric abstraction of task time. It does not capture terrain irregularity, weather, obstacle avoidance, or communication dropout, and the analytically optimal routing strategy derived in Section 3.4 is exact only for the single-headland, parallel-row field topology modeled in this paper. An irregular field with multiple access points would require a general coverage-planning or traveling-salesman-style solver rather than the closed-form solution used here.

Detection miss rates, recall-reliability values, and teleoperation latency are modeling assumptions grounded in qualitative claims drawn from the human-robot interaction and virtual reality literature reviewed in Section 2, rather than in measurements taken from an instrumented field trial. The sensitivity analysis suggests the overall conclusion is not sensitive to the exact values chosen for these parameters, but the existence and rough magnitude of the effects themselves have not been independently validated against real operating data.

The amortization model presented in Section 4.4 treats new-outbreak growth as a fixed 15 percent per-cycle rate. A validated epidemiological spread model for the specific disease in question would be a more defensible input, though the qualitative conclusion, that both agents must re-detect the field each cycle and neither receives a one-time-only detection discount, does not depend on the specific growth rate chosen.

The baseline comparison also does not account for human fatigue, error rates, or learning effects that might develop over repeated cycles, nor does it account for robot battery limitations, maintenance requirements, or failure downtime. Section 4.1 shows this is not a negligible omission: a plausible fatigue-driven speed decay over the multi-hour detection pass shrinks the human's advantage from 48.0 to 35.9 percent in the 30-target scenario, without changing its direction. Any of these factors, along with learning effects for VR-interface operators over repeated use, would affect the results reported here in practice, and a full accounting of fatigue, error, and equipment downtime for both agents remains outside the scope of this study. Most importantly, only travel time and manipulation time are modeled in this study, and treatment efficacy is assumed rather than scored: the model does not verify whether a treatment actually suppresses disease at the intended dosage or coverage, and does not score false-positive treatment of a healthy vine mistaken for a diseased one, or false-negative failure to treat a genuinely infected one beyond the single detection miss-rate parameter. Because this study evaluates time-efficiency and treatment-efficacy as independent dimensions, a faster agent that is also less thorough would appear favorably here despite a worse actual disease-management outcome; a complete evaluation of these tasks would need an accuracy- and efficacy-adjusted outcome measure in addition to a time-based one, and this is the single largest open question this study does not resolve.

Section 4.6's economic framing is deliberately incomplete: it establishes a real, sourced labor-cost figure but does not estimate a comparable robotic operating cost, since no public source gives a defensible capital and maintenance cost for this specific platform configuration. A full cost-benefit or breakeven analysis, combining this study's time results with real platform acquisition, amortization, and maintenance costs, is an important direction for future work rather than a result this paper delivers.

The optimal-routing derivation in Section 3.4 depends on headland access being available at both ends of the row block; this is standard for vineyard layout but not universal, and a field with turning space at only one end, or with irregular access points, would need a different, more general routing analysis. The empirically grounded parameters, the 24- and 50-second robot treatment dwell times, come from laboratory measurement of a specific manipulator and platform (Udekwe & Seyyedhasani, 2025a) and should not be assumed to generalize to other arm or vehicle hardware without re-measurement. Similarly, the robot speed comparison in this study is benchmarked against one specific platform's rated maximum; faster UGV platforms already exist commercially, and Section 4.3's sensitivity analysis indicates that a platform closer to human walking speed would substantially close the gap reported here.

Future work should prioritize validating the detection-reliability and teleoperation-latency parameters against instrumented field trials, extending the optimal-routing derivation to irregular field topologies using general coverage-planning methods, and evaluating the non-time-based grounds for robotic deployment, including endurance, labor availability, and treatment consistency, that this time-and-motion framework was not built to assess.

## Declaration of Competing Interests

The authors declare that they have no known competing financial interests or personal relationships that could have appeared to influence the work reported in this paper.

## Data and code availability:

The simulation code and data-generation scripts used in this study are available from the corresponding author upon reasonable request.

## CRediT authorship contribution statement

**Daniel Udekwe:** Writing – review & editing, Writing – original draft, Validation, Methodology, Investigation, Funding acquisition, Formal analysis, Data curation.
**Hassan Seyyedhasani:** Writing – review & editing, Writing – original draft, Resources.
**Muhammad Ali Qadri**: Writing – review & editing, Investigation.